\documentclass[letterpaper, 10 pt, conference]{ieeeconf}  

\IEEEoverridecommandlockouts                              

\usepackage[normalem]{ulem}	                        
\usepackage[table,usenames,dvipsnames]{xcolor}      
\usepackage{extarrows}                              
\let\labelindent\relax
\usepackage{enumitem}                               
\usepackage{amsmath,amsthm,amssymb,amsfonts,dsfont} 
\usepackage{cite}
\makeatletter
\newif\if@restonecol
\newif\foralgo@restonecol
\makeatother

\usepackage{algorithm,algorithmic}
\usepackage{graphicx}
\usepackage{subfigure}
\usepackage{tabularx}
\usepackage{booktabs}
\usepackage{multirow,multicol, array}
\usepackage[font={small}]{caption} 
\usepackage{todonotes}
\usepackage[titletoc,title]{appendix}
\usepackage[breaklinks=true, colorlinks, bookmarks=true, citecolor=Black, urlcolor=Violet,linkcolor=Black]{hyperref}
\usepackage{cleveref}
\usepackage{authblk}
\usepackage{setspace}
\usepackage{tikz}
\usepackage{float}
\usetikzlibrary{positioning, fit, shapes.geometric, arrows.meta}
\theoremstyle{definition}

\newtheorem*{assumption*}{Assumption}

\newtheorem*{remark*}{Remark}
\newtheorem*{problem*}{Problem}

\title{\LARGE \bf 
Enabling On-Chip High-Frequency Adaptive Linear Optimal Control via Linearized Gaussian Process
}

\author{Yuan Gao, Yinyi Lai, Jun Wang, Yini Fang
\thanks{Yuan Gao is with Department of Electrical and Electronic Engineering, Nanyang Technological University. {\tt\small GAOY0058@e.ntu.edu.sg}}%
\thanks{Yinyi Lai is with Department of Mechanical Engineering, Hohai University. {\tt\small 2120010118@hhu.edu.cn}}%
\thanks{Jun Wang is with Department of Electrical and Electronic Engineering, Nanyang Technological University. {\tt\small wa0004un@e.ntu.edu.sg}}%
\thanks{Yini Fang is with Department of Electronic \& Computer Engineering, Hong Kong University of Science and Technology.  
            {\tt\small yfangba@connect.ust.hk}}
\thanks{All correspondence should be sent to Yini Fang.}
}

\begin{document}

\maketitle

\begin{abstract}

Unpredictable and complex aerodynamic effects pose significant challenges to achieving precise flight control, such as the downwash effect from upper vehicles to lower ones. Conventional methods often struggle to accurately model these interactions, leading to controllers that require large safety margins between vehicles. Moreover, the controller on real drones usually requires high-frequency and has limited on-chip computation, making the adaptive control design more difficult to implement.
To address these challenges, we incorporate Gaussian process (GP) to model the adaptive external aerodynamics with linear model predictive control. The GP is linearized to enable real-time high-frequency solutions. Moreover, to handle the error caused by linearization, we integrate end-to-end Bayesian optimization during sample collection stages to improve the control performance. Experimental results on both simulations and real quadrotors show that we can achieve real-time solvable computation speed with acceptable tracking errors.

\end{abstract}

\section{INTRODUCTION}

The growing commercial availability of unmanned aerial vehicles (UAVs) is driving interest in sophisticated control techniques for large-scale aerial swarms \cite{abdelkader2021aerial,chung2018survey,morgan2016swarm}. These methods are applicable in various settings including manipulation, search, surveillance, and mapping. Such environments often necessitate that UAVs maintain minimal distances between each other, a strategy referred to as dense formation control. For instance, in a search-and-rescue operation within a collapsed structure, the ability of the swarm to operate in close proximity allows for faster navigation through the building compared to swarms required to keep larger separations.

A significant challenge in close-proximity control is the intricate aerodynamic interactions caused by the minimal distances maintained between UAVs. For example, when one multirotor hovers above another, it induces a phenomenon known as the downwash effect on the drone below, which poses challenges for traditional modeling techniques\cite{danjun2015autonomous,kan2019analysis}. In the absence of advanced models for these interactions, it becomes necessary to enforce a substantial safety margin between the UAVs, such as 60cm for the compact Crazyflie 2.0 quadrotor, which has a rotor-to-rotor span of 9cm\cite{honig2018trajectory}. Nevertheless, when two Crazyflie quadrotors are positioned just 30cm apart vertically, the resulting downwash measures only -9g, a level comfortably within their thrust limits. This observation underscores that refined modeling of downwash and other aerodynamic interactions could substantially enhance precision in dense formation control.

 Model Predictive Control (MPC) facilitates precise adaptive control for robots, ensuring compliance with intricate control and state constraints, particularly in scenarios involving dynamic obstacle avoidance and contact events\cite{wensing2023optimization,di2020software,manchester2019contact}. However, for the close proximal drone control, we usually do not know the exact aerodynamic model of downwash effect, making it difficult for MPC to predict and safely stabilize the drone, where data-driven model is necessary.

The Gaussian process is a commonly utilized data-driven compensatory mechanism within model-based controllers, as it demonstrates efficiency in sampling and operates effectively without prior model knowledge \cite{xu2014gp, wilcox2020solar}. When integrated with MPC, the Gaussian process can provide an accurate and uncertainty-aware model of the system \cite{liu2018gaussian, schmid2022real, zheng2022gp}, thereby enabling the MPC to adapt its predictions in response to external disturbances. Nonetheless, data-driven models typically entail substantial computational overhead. In scenarios characterized by constrained computational resources such as drone navigation, the integration of nonlinear GP with MPC becomes unfeasible. Earlier investigations \cite{nghiem2019linearized, nghiem2019fast} have demonstrated the potential of Linearized Gaussian Processes (LinGP). However, the direct implementation of LinGP on actual drone systems is still too heavy for real-time computations.

Recent work has shown that linear MPC can be made on-chip computable on the Crazyflie \cite{nguyen2024tinympc}. To this end, we propose a novel approach to incorporate GP into linear MPC to adaptively predict the surrounding aerodynamics for homogeneous swarms of multirotor UAVs. Our approach employs a GP to estimate interaction forces among multirotors that are typically unaccounted for by standard free-space aerodynamic models. Then the GP is linearized and combined with linear MPC to achieve real-time high-frequency control. To compensate for the control error caused by linearization, we utilize the end-to-end Bayesian optimization approach to collect samples based on the performance of our linearized adaptive MPC controller. Experimental results on both simulation and real quadrotors show that the proposed solution demonstrates good computational efficiency, allowing for real-time implementation on compact 32-bit microcontrollers, and good control performance in resisting downwash effects.

Our contributions are threefold:
\begin{enumerate}
 \item Incorporate linearized GP to make the linear MPC both adaptive to external disturbance and computable on real-time embedded microcontrollers.
    \item To handle the error caused by linearization, we use end-to-end Bayesian optimization with the additive Gaussian process that leverages the previous force prediction results.
    \item Simulation and real drone verify the effectiveness of the proposed controller on mitigating the downwash effects.
\end{enumerate}

\section{Related Works}
Several machine learning techniques have been effectively integrated with Model Predictive Control (MPC) to enhance its performance and extend its capabilities. Notably, Gaussian Process (GP) regression has been extensively utilized to automatically learn unmodeled (often nonlinear) dynamics within the prediction model\cite{hewing2019cautious}, and to construct provably accurate confidence intervals for predicted trajectories\cite{koller2018learning}. 

Building upon the approaches in \cite{hewing2019cautious} and \cite{kabzan2019learning}, we employ Gaussian Processes to capture the residual dynamics relative to a simplified quadrotor model that omits aerodynamic effects. By focusing on residual dynamics, we reduce the complexity of the learning task, which enables the effective augmentation of the model with just a few inducing points for Gaussian Processes. This compact representation facilitates the integration of the enhanced dynamics model within a Model Predictive Control (MPC) framework.

In terms of motion planning, empirical models have been utilized to prevent adverse interactions\cite{morgan2014model,debord2018trajectory}. These models typically represent safe interaction regions as ellipsoids or cylinders, which are effective for both homogeneous and heterogeneous multirotor teams. However, determining these shapes often requires risky flight tests and results in conservative estimates. In contrast, we use learning to more accurately estimate interaction forces and integrate these forces into the controller to enhance trajectory tracking during close-proximity flight. The estimated forces could also be potentially applied to improve motion planning.

\section{Problem Statement}
\subsection{Linearized dynamics of the quadrotors}

Given quadrotor states as global position $\mathbf{p} = [p_x, p_y, p_z] \in \mathbb{R}^3$, velocity $\mathbf{v} = [ v_x, v_y, v_z ] \in \mathbb{R}^3$, attitude rotation matrix $R \in \mathrm{SO}(3)$, and body angular velocity $\omega = [\omega_x, \omega_y, \omega_z]\in \mathbb{R}^3$, we consider the following dynamics:

\begin{equation}
\label{eq:dynamics of the quadrotors}
\begin{array}{ll}
\dot{\mathbf{p}}=\mathbf{v}, & m \dot{\mathbf{v}}=m \mathbf{g}+R \mathbf{f}_u+\mathbf{f}_a \\
\dot{R}=R S(\boldsymbol{\omega}), & J \dot{\boldsymbol{\omega}}=J \boldsymbol{\omega} \times \boldsymbol{\omega}+\boldsymbol{\tau}_u+\tau_a
\end{array}
\end{equation}

where $m$ and $J$ are mass and inertia matrix of the system respectively, $S(\cdot)$ is skew-symmetric mapping. $\mathbf{g}=[0,0,-g]^{\top}$ is the gravity vector, $\mathrm{f}_u=[0,0, T]^{\top}$ and $\tau_u=\left[\tau_x, \tau_y, \tau_z\right]^{\top}$ are the total thrust and body torques from four rotors predicted by a nominal model. We use $\boldsymbol{\eta}=\left[T, \tau_x, \tau_y, \tau_z\right]^{\top}$ to denote the output wrench. Typical quadrotor control input uses squared motor speeds $\mathbf{u}=\left[n_1^2, n_2^2, n_3^2, n_4^2\right]^{\top}$, and is linearly related to the output wrench.
We linearized the dynamics near the hovering position and stable hovering inputs $u_{\text{hover}}$. The quadrotor states are defined as $x=\left[p_x, p_y, p_z, \theta_r, \theta_p, \theta_y, v_x, v_y, v_z, \omega_x, \omega_y, \omega_z\right]^T$, which includes the position, rotation angle, velocity and angular velocity of the quadrotor. 

\subsection{Dynamic Behavior of Swarm Systems}
Assume that we are working with \(n\) homogeneous multirotor systems. For simplicity, the state of the \(i\)-th multirotor can be denoted as \(\mathbf{x}^{(i)} = \left[\mathbf{p}^{(i)}; \mathbf{v}^{(i)}; \mathbf{R}^{(i)}; \boldsymbol{\omega}^{(i)}\right]\), where \(\mathbf{p}^{(i)}\) represents the position, \(\mathbf{v}^{(i)}\) is the velocity, \(\mathbf{R}^{(i)}\) is the rotation matrix, and \(\boldsymbol{\omega}^{(i)}\) is the angular velocity. Equation \eqref{eq:dynamics of the quadrotors} can then be rewritten in a simplified form as:
\begin{equation}
\dot{\mathbf{x}}^{(i)} = f(\mathbf{x}^{(i)}, \mathbf{u}^{(i)}) + 
\begin{bmatrix}
0 \\
\mathbf{f}_a^{(i)} \\
0 \\
\boldsymbol{\tau}_a^{(i)}
\end{bmatrix}, \quad i = 1, 2, \dots, n,
\end{equation}
where \(\mathbf{f}_a^{(i)}\) and \(\boldsymbol{\tau}_a^{(i)}\) represent the external forces and torques applied to the \(i\)-th multirotor respectively.

\subsection{Problem Statement and Proposed Approach}
Our goal is to enhance the control performance of a multirotor swarm when flying in close formation by identifying the unknown interaction terms \(\mathbf{f}_a\) and \(\boldsymbol{\tau}_a\). In this section, we mainly address the position dynamics, where \(\mathbf{f}_a\) plays a crucial role.

To achieve this, we begin by approximating \(\mathbf{f}_a\) using Gaussian
Process (GP)  that maintains permutation invariance. This GP model is then integrated into our controller, which ensures exponential stabilization. 

\begin{figure}
    \centering
    \includegraphics[width=0.95\linewidth]{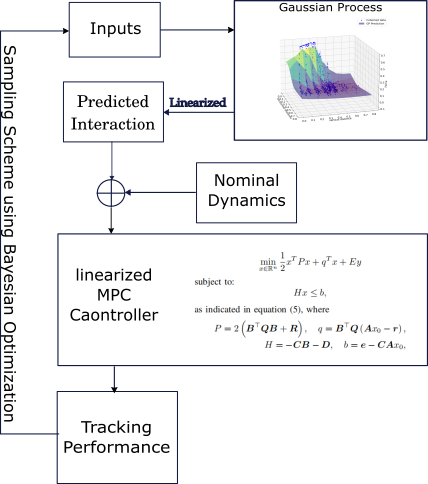}
    \caption{Exploring complex interactions between multirotors, we developed a GP-based controller for close-proximity flight, using the position difference between the two drones as input.
}
    \label{fig:enter-label}
\end{figure}

\section{Preliminaries}

\subsection{Gaussian Process and Bayesian Optimization}
We introduce the Gaussian process (GP), a probabilistic framework utilized for modeling the objective function. The GP is constructed on observed data \(D_t\) and relies on a positive-definite kernel function \(k(x, x')\), which embodies our prior understanding of the function \(f(x)\) and its correlation with \(f(x')\). With \(D_t\) and \(k\), the GP framework fully characterizes the function \(f\) through its mean function \(\mu_t(x)\) and covariance function \(\Sigma_t(x, x')\), computed as follows:
\begin{equation}
\begin{aligned}
\mu_t(x) &= k_t(x)^T (K_t + \sigma^2 I)^{-1} y_t, \\
\Sigma_t(x, x') &= k(x, x') - k_t(x)^T (K_t + \sigma^2 I)^{-1} k_t(x').
\end{aligned}\label{eq:gp_regression}
\end{equation}

Let $k_t(x) = [k(x', x)]_{x' \in \mathcal{X}_{t-1}}^\top$ and $y_t = [y]_{y \in \mathcal{Y}_{t-1}}^\top$ represent $m \cdot (t-1)$ dimensional vectors. The kernel matrix $K_t = [k(x, x')]_{x, x' \in \mathcal{X}_{t-1}}$ is a positive definite matrix of size $m \cdot (t-1) \times m \cdot (t-1)$. Here, $I$ denotes the identity matrix with dimensions matching $K_t$. The Gaussian Process (GP) model is then represented as $GP(\mu_t, \Sigma_t \mid D_t)$. We define the variance function $\sigma_t^2(x) = \Sigma_t(x, x)$ and set $\sigma_t(x) = \sqrt{\sigma_t^2(x)}$. The term $\mu_t(x)$ represents the mean prediction of the function $f(x)$ based on the observed data $D_t$, while $\sigma_t^2(x)$ captures the uncertainty in this prediction, expressed as variance\cite{williams2006gaussian,ma2023gaussian}. 

\textbf{Additive Gaussian processes} (add-GP) were proposed in \cite{duvenaud2011additive}, and analyzed in the BO setting in \cite{kandasamy2015high}. Following the latter, we assume that the function $f$ is a sum of independent functions sampled from Gaussian processes that are active on disjoint sets $A_m$ of input dimensions. Precisely, $f(x)=\sum_{m=1}^M f^{(m)}\left(x^{A_m}\right)$, with $A_i \cap A_j=$ $\emptyset$ for all $i \neq j,\left|\cup_{i=1}^M A_i\right|=d$, and $f^{(m)}$ $G P\left(\mu^{(m)}, k^{(m)}\right)$, for all $m \leq M(M \leq d<\infty)$. As a result of this decomposition, the function $f$ is distributed according to $G P\left(\sum_{m=1}^M \mu^{(m)}, \sum_{m=1}^M k^{(m)}\right)$. Given a set of noisy observations $D_t=\left\{\left(\boldsymbol{x}_\tau, y_\tau\right)\right\}_{\tau=1}^t$ where $y_\tau \sim \mathcal{N}\left(f\left(x_\tau\right), \sigma^2\right)$, the posterior mean and covariance of the function component $f^{(m)}$ can be inferred as $\mu_t^{(m)}(\boldsymbol{x})=\boldsymbol{k}_t^{(m)}(\boldsymbol{x})^{\mathrm{T}}\left(\boldsymbol{K}_t+\sigma^2 \boldsymbol{I}\right)^{-1} \boldsymbol{y}_t$ and $k_t^{(m)}\left(\boldsymbol{x}, \boldsymbol{x}^{\prime}\right)=k^{(m)}\left(\boldsymbol{x}, \boldsymbol{x}^{\prime}\right)-\boldsymbol{k}_t^{(m)}(\boldsymbol{x})^{\mathrm{T}}\left(\boldsymbol{K}_t+\right.$ $\left.\sigma^2 \boldsymbol{I}\right)^{-1} \boldsymbol{k}_t^{(m)}\left(\boldsymbol{x}^{\prime}\right)$, where $\boldsymbol{k}_t^{(m)}(\boldsymbol{x})=\left[k^{(m)}\left(\boldsymbol{x}_i, \boldsymbol{x}\right)\right]_{\boldsymbol{x}_i \in D_t}$ and $\boldsymbol{K}_t=\left[\sum_{m=1}^M k^{(m)}\left(\boldsymbol{x}_i, \boldsymbol{x}_j\right)\right]_{\boldsymbol{x}_i, \boldsymbol{x}_j \in D_t}$. For simplicity, we use the shorthand $k^{(m)}\left(\boldsymbol{x}, \boldsymbol{x}^{\prime}\right)=k^{(m)}\left(\boldsymbol{x}^{A_m}, \boldsymbol{x}^{A_m}\right)$.

\subsection{Linear-Quadratic Regulator and Linear MPC}
\label{sec:linear_mpc}
The Linear-Quadratic Regulator (LQR) is a well-established method used for tackling control issues in robotics. LQR seeks to minimize a quadratic cost function constrained by linear dynamics equations. The problem is formulated as follows:

\begin{equation}
\begin{aligned}
\min_{x_{1:N}, u_{1:N-1}} J &= \frac{1}{2} x_N^\top Q_f x_N + q_f^\top x_N + \\
&\quad \sum_{k=1}^{N-1} \left( \frac{1}{2} x_k^\top Q x_k + q_k^\top x_k + \frac{1}{2} u_k^\top R u_k + r_k^\top u_k \right) 
\end{aligned}
\end{equation}

subject to the linear system dynamics:
\begin{equation}
x_{k+1} = A x_k + B u_k, \quad \forall k \in [1, N), 
\end{equation}

where $x_k \in \mathbb{R}^n$ and $u_k \in \mathbb{R}^m$ represent the state and control input at step $k$, respectively. The matrices $A \in \mathbb{R}^{n \times n}$ and $B \in \mathbb{R}^{n \times m}$ define the system dynamics, $Q, Q_f \geq 0$, and $R > 0$ are symmetric cost matrices, and $q, q_f, r$ are linear cost vectors.

Equation \eqref{eq:dynamics of the quadrotors} admits an analytic solution in the form of an affine feedback controller:
\begin{equation}
u_k = -K_k x_k - d_k,  
\end{equation}
where the feedback gains $K_k$ and feedforward terms $d_k$ are derived from solving the backward recursive discrete-time Riccati equations \cite{lewis2012optimal,kirk2004optimal,todorov2006optimal}. Linear MPC expands upon the LQR by incorporating reference tracking and additional convex constraints that pertain to both system states and control inputs, we will show our problem formulation in the next section.

\section{controller design}
We design a controller that combines Linearized Gaussian Processes (LinGP) with Adaptive Linear MPC(LinMPC) to adaptively manage external disturbances during the flight of
multiple UAVs. This approach utilizes a dynamic model to
enhance performance, ensuring real-time solvability.

Additionally, to address the control errors introduced by linearization, we employ an end-to-end Bayesian optimization method to gather samples, improving the performance of our linearized adaptive MPC controller.

\subsection{Linearized Gaussian Processes}
We begin with how to linearized the Gaussian process assuming we already have a dataset of input $x$ and measurement $y$ here. The input $x$ in our case is the 3-dimensional position difference of two drones, and output $y$ is the measured downwash force. After collecting the dataset, we directly utilize the mean value as the predicted force, thus

\begin{equation}
\begin{aligned}
    \mathbf{f}_u(d) &= k_t(d)^T (K_t + \sigma^2 I)^{-1} y_t \\
    &:= k_t(d)^T w \approx k_t(d_0)^T w + (d-d_0)^\top \nabla_d k(d_0) ^\top w \\ 
\end{aligned}
\end{equation}
where $w = (K_t + \sigma^2 I)^{-1} y_t$ is fixed after data collection, $d$ is the input, i.e., the drone position difference. The $\approx$ simply comes from the ignoring high-order terms in first-order Taylor expansion.

Appending the linearized aerodynamics model to the linearized drone model, we have the abstract version of drone dynamics model,

\begin{equation}
    x' = Ax + Bu + Dd+c\label{eq:dynamics_with_disturbance}
\end{equation}
where $c=k_t(d_0)^T w - \nabla_d k(d_0)^\top $. $Dd\:=\Delta x$ is the external force resulting from downwash. The formulation of $D$ is obtained through the linearization of the Gaussian Process (GP). The prediction model of relative distance:
\begin{equation}
    \dot \Delta x = \Delta v\label{eq:delta_x_predction}
\end{equation}

\subsection{Adaptive Linear MPC}

Compared to the linear MPC shown in Section, we add constraints including collision avoidance constraints and actuation limits:

\begin{equation}
\min_{x_{1:N}, u_{1:N-1}} J(x_{1:N}, u_{1:N-1})
\end{equation}
subject to the system dynamics:

\begin{equation}
\begin{aligned}
x_{k+1} = A x_k + B u_k + Dd, \quad \forall k \in [1, N), \\
x_k \in \mathcal{X}, \quad u_k \in \mathcal{U},\label{eq:the system dynamics11}
\end{aligned}
\end{equation}
where $\mathcal{X}$ and $\mathcal{U}$ are polytope safe and acturation limit sets. 
Because the constraints $\mathcal{X}$ and $\mathcal{U}$ are linear, the problem described by equation \eqref{eq:the system dynamics11} becomes a quadratic program (QP) and can be formatted as follows:
\begin{equation}
\min_{x \in \mathbb{R}^n} \frac{1}{2} x^T P x + q^T x + Ey\label{eq:controller}
\end{equation}
subject to:
\begin{equation}
H x \leq b,
\end{equation}
as indicated in equation \eqref{eq:controller}, where 
$$
\begin{aligned}
    P=2\left(\boldsymbol{B}^{\top} \boldsymbol{Q} \boldsymbol{B}+\boldsymbol{R}\right), \quad q=\boldsymbol{B}^{\top} \boldsymbol{Q}\left(\boldsymbol{A} x_0-\boldsymbol{r}\right), \\
    \quad H=-\boldsymbol{C} \boldsymbol{B}-\boldsymbol{D}, \quad b=\boldsymbol{e}-\boldsymbol{C} \boldsymbol{A} x_0,
\end{aligned}
$$
and
{\small
$$
\boldsymbol{A}=\left[\begin{array}{c}
A \\
A^2 \\
\vdots \\
A^N
\end{array}\right], \boldsymbol{B}=\left[\begin{array}{cccc}
B & & & \\
A B & B & & \\
\vdots & & \ddots & \\
A^{N-1} B & \cdots & & B
\end{array}\right],
$$
}
$$
\begin{aligned}
& \boldsymbol{C}=I_N \otimes\left[I_{n_{\text {sys }}}-I_{n_{\text {sys }}} \mathbf{0}_{n_{\text {sys }} \times 2 m_{\text {sys }}}\right]^{\top} \\
& \boldsymbol{D}=I_N \otimes\left[\mathbf{0}_{m_{\text {sys }} \times 2 n_{\text {sys }}} I_{m_{\text {sys }}}-I_{m_{\text {sys }}}\right]^{\top} \\
& \boldsymbol{e}=I_N \otimes\left[x_{\max }^{\top}-x_{\min }^{\top} u_{\max }^{\top}-u_{\min }^{\top}\right]^{\top} \\
& \boldsymbol{r}=I_N \otimes r, \boldsymbol{Q}=I_N  \otimes Q, \boldsymbol{R}=I_N \otimes R\\
& E = \left[A^{n-1}D,\dots, D\right], y = [d_0,d_1
,\dots, d_{N-1}]^\top
\end{aligned}
$$

The problem is still a QP problem\cite{lu2024mpc}, where we can call an arbitrary QP solver to solve it. In practice, we use \cite{nguyen2024tinympc} for better compatibility with drones.

\section{Sampling Scheme using Bayesian Optimization}
While our linearization scheme is very naive, one may question how the performance must be sacrificed due to our linearization. While for real-time solutions on the drones, we must trade off the model preciseness and computation burden. However, we tried to use the sampling scheme to make up the tracking performance by choosing the sample trajectories. Therefore, we introduce the sampling scheme based on Bayesian optimization.

\noindent\textbf{Bayesian Optimization with additive Gaussian process.} Bayesian optimization (BO) is a sample-efficient method for maximizing expensive-to-evaluate black-box functions, which frequently arise in robotic applications. By iteratively evaluating the black-box function at the \emph{query points}, BO first builds a probabilistic model about the function and then infers the location of the maximum accordingly. 

The most commonly algorithm for query point selection include GP-UCB \cite{srinivas2009gaussian}, which selects queries by the maximum point of the posterior mean value plus a scaled posterior standard deviation,
$
x = \max_{x\in \mathcal{X}}\mu_t(x) + \beta_t\sigma_t(x)
$,
where $\beta_t$ is the confidential weights interval hyperparameters. 

For better improve performance via adapting the sampling process, we can set the tracking performance as our ultimate objective to guide the sampling points of our GP model. Moreover, to better leverage the GP downforce model we have previously, we leverage the additive GP to add another GP on the downforce GP to model the end-to-end performance from the sampling point to tracking performance.

One problem in the sampling scheme is that we can sample many points on a single trajectory for the downwash force prediction, while we only have one measure of the overall performance, which is not consistent. As our ultimate goal is to optimize the performance and the two drones will follow specific structure given , we sub-sample the whole trajectories by the most informative point to performance GP, and use this point to guide the sampling scheme. More specifically, we follow the following sampling scheme in Algorithm \ref{alg:Bayesian Optimization Guided Sample Scheme of Adaptive Linear MPC}.

\begin{algorithm}[H]
    \caption{Bayesian Optimization Guided Sample Scheme of Adaptive Linear MPC}
    \label{alg:Bayesian Optimization Guided Sample Scheme of Adaptive Linear MPC}
    \begingroup
    \small 
    \begin{algorithmic}[1]
        \REQUIRE Upper drones $D_1$, lower drones (with MPC controller) $d_2$, Initial sample dataset for down wash prediction $\mathcal D^{\rm Force}=\emptyset$, sample dataset for performance prediction $D^{\rm Force}=\emptyset$, downwash force prediction GP $f\sim GP^{\rm Force}(\mu^{\rm Force}_0,\Sigma^{\rm Force}_0)$, performance prediction GP $GP^{\rm Perf}(\mu^{\rm Perf}_0,\Sigma^{\rm Perf}_0)$, Given reference tracking trajectory $\mathcal T$, Drone interaction region $\mathcal N\in\mathcal{X}$
        \FOR{sample time step $t=1,2,\dots,T$}
        \STATE Generate interaction sample target $x=(\Delta x, \Delta y, \Delta z)$ based on GP prediction on additive GP, 
        $$
        x = \arg\max_{x\in\mathcal \mathcal N} \mu^*_{t-1}(x) + \beta_t \sigma^{\rm Perf}_{t-1}(x)
        $$
        where $\mu^*(x) = \mu^{\rm Perf} + \mu^{\rm Force} $ and $\sigma^* = \sigma^{\rm Perf}+\sigma^{\rm Force}$ is from the added GP, only based on performance prediction dataset $\mathcal{D}^{\rm Perf}$.
        \STATE Call the motion planning oracle to plan trajectories for both two drones
        \STATE Predict down wash force model based on current $GP_{\rm Force} $ and input to linMPC-linGP controller \eqref{eq:controller} on $d_2$ 
        \STATE Uses the controller to track the target trajectories, measure force data and track error 
        \STATE Append all interaction data in to force prediction dataset,
        $$
        D^{\rm Force}_t = D^{\rm Force}_{t-1}\cup\{(x, f(x) \mid \forall x \in \mathcal N\}
        $$
        \STATE Append most informative point in the interaction point in the performance prediction dataset,
        $$
        D^{\rm Perf}_t = D^{\rm Perf}_{t-1}\cup\{(x, J(x) \mid x = \arg\max_{x\in\mathcal N} \sigma^{\rm Perf}(x)\}
        $$
        \STATE Apdate both GP with datasets $D^{\rm Force}_t$ and $D^{\rm Perf}_t$
        \ENDFOR
    \end{algorithmic}
    \endgroup
\end{algorithm}

\begin{table*}[h]

\caption{Comparsion of average tracking error in simulation. The proposed LinMPC-LinGP has comparable tracking error while significantly lower computation time compared to other nonlinear MPC baselines, while the proposed Bayesian optimization sampling scheme further reduce the tracking error. } \label{table}
\centering
\begin{tabular}{@{}cccccc@{}}
\toprule
&\multicolumn{4}{c}{Tracking performance } &Computation\\
  Tracking performance                          & $\Delta d=0.5$ (m) & $\Delta d=0.4$ (m) & $\Delta d=0.3$ (m) & $\Delta d=0.2$ (m) &  Time (s) \\ \midrule
LinMPC-LinGP+BO (ours)          & 0.0026              & 0.014              & 0.471              & 1.081              & 0.009               \\
LinMPC-LinGP (ours)          & 0.0034              & 0.020              & 0.528              & 1.432              & 0.009               \\
PID                          & 0.34               & 0.084              & 2.94               & 8.827              & 0.0002              \\
MPC-LinGP                    & 0.0039             & 0.011              & 0.343              & 0.914              & 0.3                 \\
MPC-FullGP                   & 0.0020             & 0.010              & 0.279              & 0.766              & 0.87                \\ \bottomrule
\end{tabular}
\end{table*}

\section{Experimental Results}
\label{sec:numerical}

We design our evaluation procedure to address the following questions:  (1) What is the contribution of the learned dynamics of our GP-MPC in a downwash task? (2) Is the linearized GP effective in predicting the downwash effect? How is the performance and computation compared to full GP? (3) When we transfer the algorithm that combines GP and MPC to a real robot, how does the algorithm perform?
\begin{figure*}[h]
    \centering
    \includegraphics[width=\linewidth]{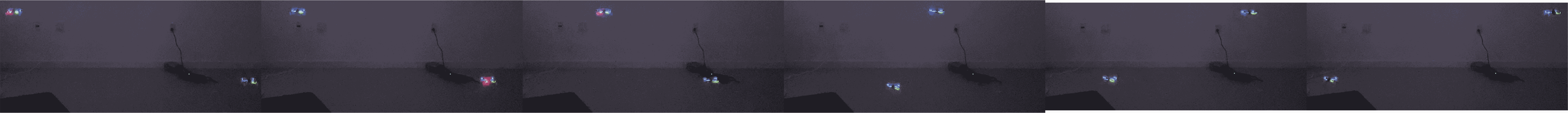}
    \caption{Screenshots of LinGP-LinMPC controller on real drones, which is able to predict the downwash force and keep stable.}
    \label{fig:mpc}
    \includegraphics[width=\linewidth]{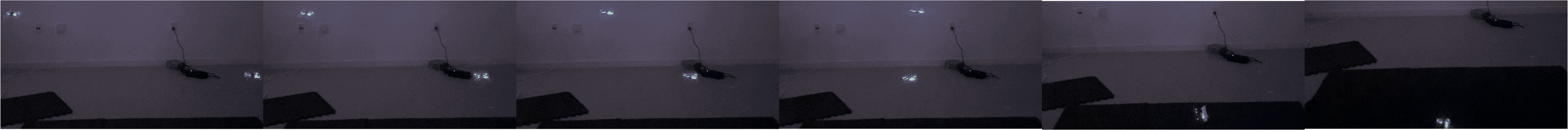}
    \caption{Screenshots of built-in Mellinger controller on real drones, which is not able to predict keep stable and crash.}
    \label{fig:builtin}
\end{figure*}
\subsection{Simulation}
\subsubsection{Experiment Setup}
\begin{figure}[H]
    \centering
    \includegraphics[width=0.8\linewidth]{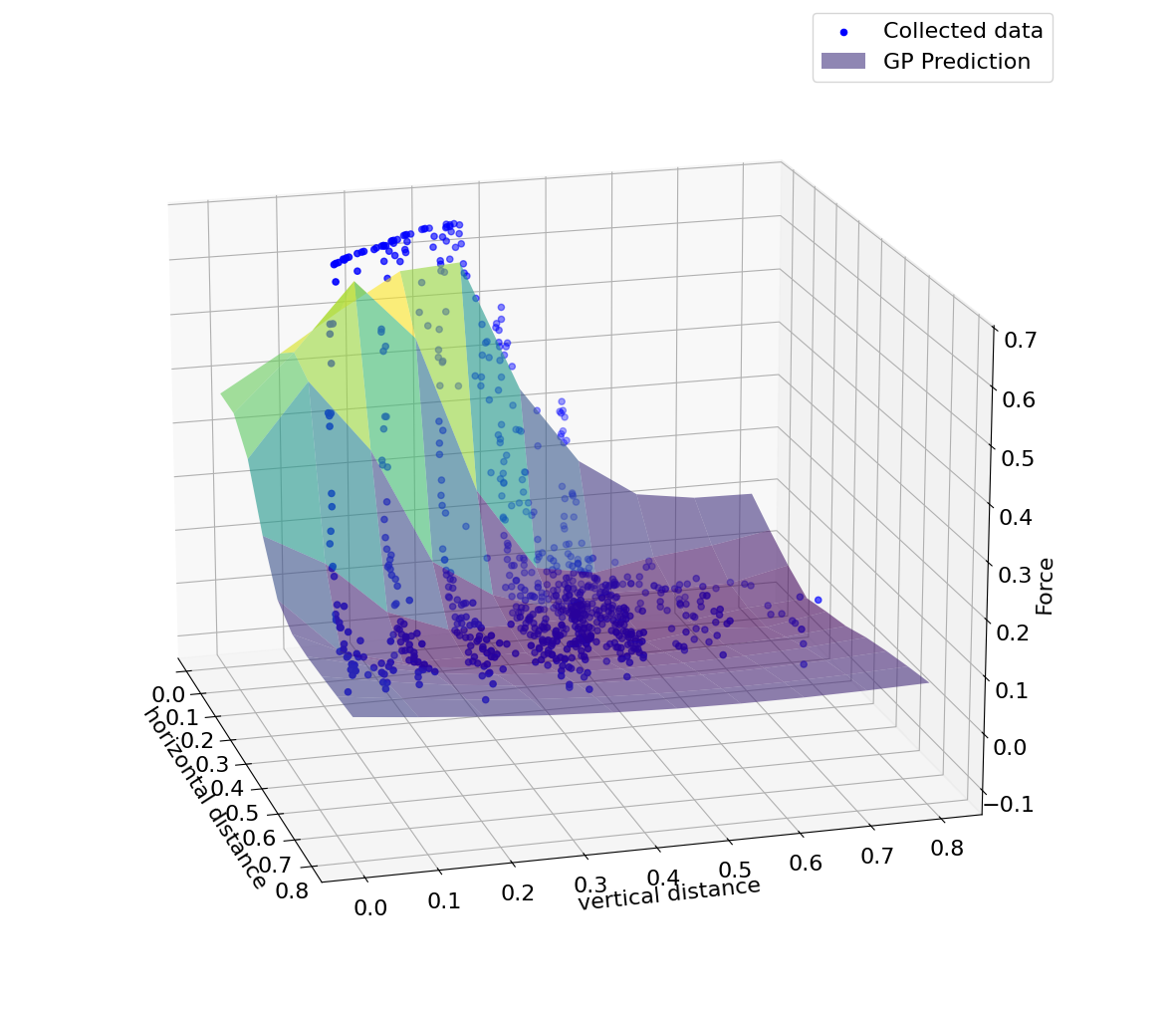} 
    \caption{The Gaussian process model is used to capture the main trends of the data and make force predictions based on the changes in the horizontal and vertical distances.}
    \label{fig:force_prediction}
\end{figure}
\begin{figure}[h]
    \centering
    \includegraphics[width=1.0\linewidth]{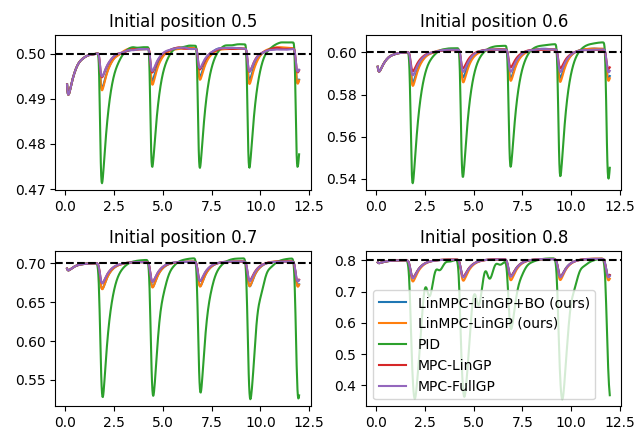}
    \caption{The tracking performance of various controllers on a swapping task across different initial positions in the simulator.}
    \label{fig:tracking}
\end{figure}
We utilized the PyBullet Physics Simulator as the simulation platform\cite{panerati2021learning}, along with the TinyMPC package for dynamic modeling\cite{nguyen2024tinympc} and a Gaussian Process (GP) package for disturbance modeling. The solver integrated within the platform ensured accurate and efficient computation for our control algorithms.

The experiments were designed to assess the performance of our controller by simulating two drones flying towards each other at different heights. We specifically evaluated whether the controller reduced the tracking error in the z-axis during the flight.

\subsubsection{Experimental Results}
To determine the disturbance force $f_a$, a skilled pilot manually flew the UAV at varying altitudes, and training data was collected consisting of sequences of state estimates and control inputs $\{(\mathbf{p}, \mathbf{v}, \mathbf{R}, \mathbf{u}), y\}$, where $y$ is the measured value of $f_a$. The relationship $f_a = m\dot{\mathbf{v}} - mg - Rf_u$ was used to compute $f_a$ (from equation \eqref{eq:dynamics of the quadrotors}). 

In this study, we focus solely on learning the \(z\)-axis component of \(\mathbf{f}_a\), as our analysis indicates that the \(x\) and \(y\) components are negligible and have minimal impact on the overall system dynamics.

\textbf{Gaussian process model learning.}
We visualize the Gaussian process learning results in Fig. \ref{fig:force_prediction}. We use a rational quadratic kernel. The graph shows the predicted downwash force, where the blue dots represent the actual collected data and the purple to green gradient surfaces represent the results predicted by the Gaussian process.

\textbf{Algorithm and Baselines}. The four controllers being compared in Fig. \ref{fig:tracking} are: 
\begin{enumerate}
    \item \textbf{LinMPC-LinGP+BO (ours)}: The proposed Adaptive linear MPC -linear GP controller with BO sample scheme.
    \item \textbf{LinMPC-LinGP (ours)}: The proposed Adaptive linear MPC -linear GP controller without BO sample scheme. The data is sampled from randomly generated initial locations.
    \item \textbf{MPC-LinGP}: This method combines MPC with a Linear GP, similar to the proposed method but differing in its internal structure. It performs better than PID, but exhibits larger oscillations compared to the proposed LinMPC-LinGP approach. 
    \item \textbf{PID}: This refers to the Proportional-Integral-Derivative controller. The performance shows periodic oscillations and less accurate tracking compared to the GP-based methods, particularly for higher initial positions.
    \item \textbf{MPC-FullGP}: This is the combination of MPC with a Full GP model. It shows the largest deviation and oscillations in tracking the target, especially at higher initial positions, indicating less robust performance in this context.
\end{enumerate}

Across all four initial positions (0.5, 0.6, 0.7, and 0.8), both two proposed LinMPC-LinGP method demonstrate the comparable accurate tracking to other two nonlinear MPC, closely following the reference trajectories (represented by the dashed line).

Table. \ref{table} compares the count of the average computation time for four different control strategies. The PID controller is the most efficient while MPC-FullGP is the most computationally demanding. The proposed LinMPC-LinGP strikes a balance between performance and computational efficiency, making it a practical choice for real-time applications.

\begin{figure}[h]
    \centering
    \includegraphics[height=0.5\linewidth]{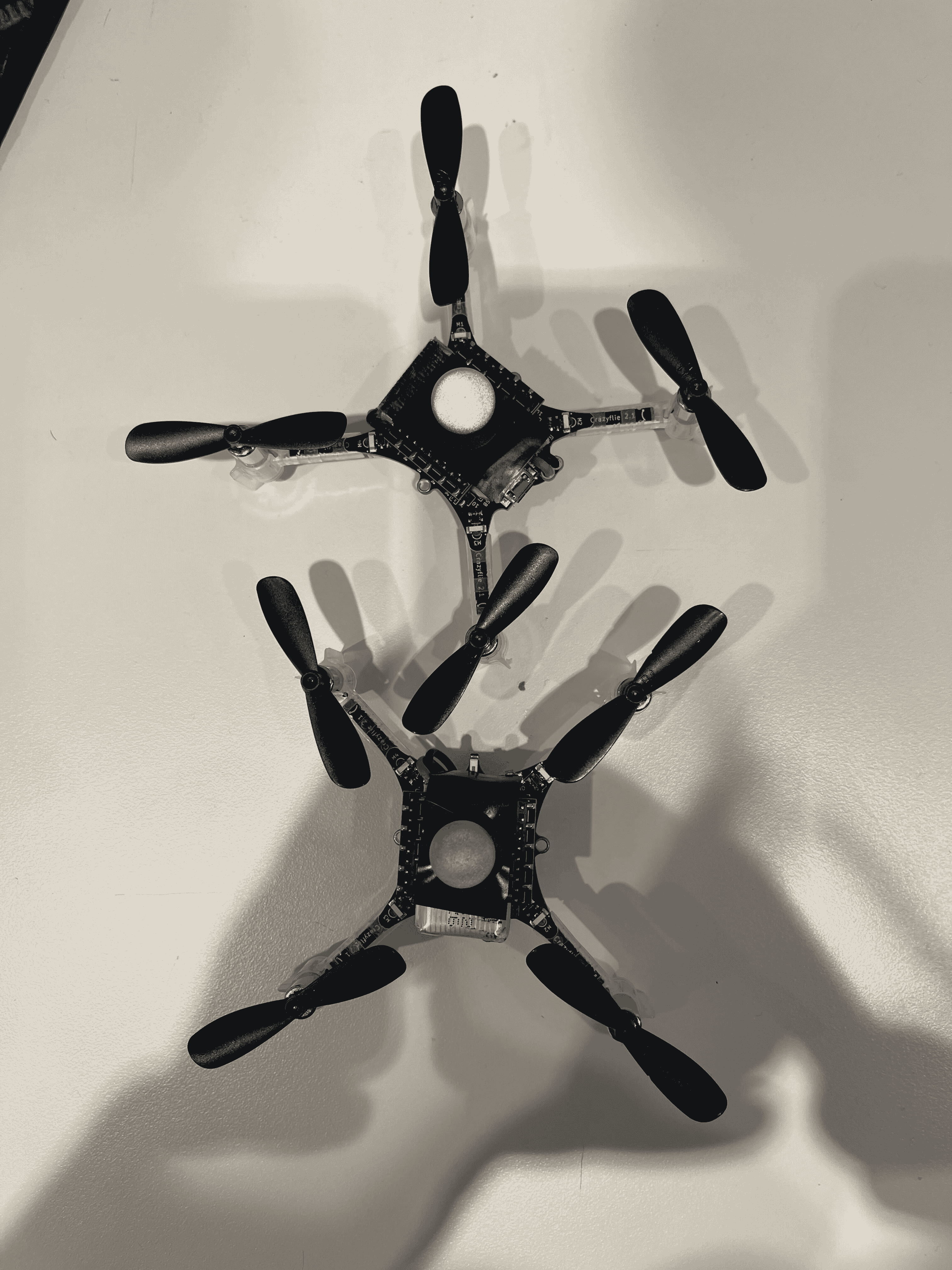}
    \includegraphics[height=0.5\linewidth]{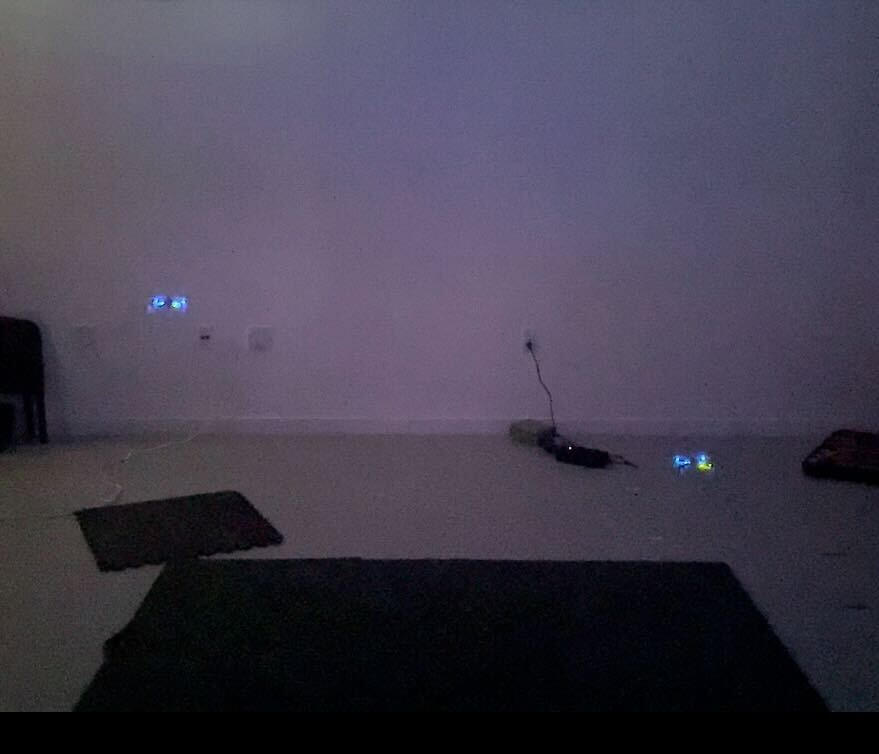}
    \caption{Crazyflie and the experiment environment.}
    \label{fig:exp_env}
\end{figure}


\subsection{Real Robot Experiment}

\subsubsection{Experiment Setup}
We show the effectiveness of our learning-based controller on the Crazyflie 2.1 Quadrotor\cite{budaciu2019evaluation} with a STM32F405 micro-controller clocked at 168 MHz shown in Figure \ref{fig:exp_env}. We  handle the communication using Crazyswarm\cite{preiss2017crazyswarm,pichierri2023crazychoir}. We use the Optitrack motion capture system to provide localization for the crazyflies.

\subsubsection{Real robot Implementations} We implement the linMPC-linGP controller based on the implementation of \cite{nguyen2024tinympc}, while we fixed the linearized drone model but only update the linearized GP model $\mathrm{D}$ as parameters in real time. To reduce communication burden via sending multiple parameters every communication step, we update the parameters $D$ every 0.1 second.

\subsubsection{Experimental Setup and Results} We ask two drones to swap their x-y positions with height difference 0.4m with in 6 seconds. We take the screen shot every one seconds and shown in Figuer \ref{fig:mpc} and \ref{fig:builtin}. The built-in PID controller fails to maintain stable, while the proposed adaptive linear MPC controller successfully keeps stable and arrives at the target.

\section{Conclusion}
In this study, we design a learning-based controller to facilitate the close-proximity flight of multiple UAVs. We introduce the application of Gaussian Processes (GPs) to model the interaction forces among several quadrotors, depending on the relative positions and velocities of adjacent aircraft. Through comprehensive experiments conducted in both simulated environments and real-world scenarios, we demonstrate that our method surpasses the performance of other control method. The proposed LinMPC-LinGP method offers an optimal trade-off between computational efficiency and system performance, making it highly suitable for deployment in real-time scenarios.

There are several potential avenues for future research. To be specific, our approach could be extended to accommodate heterogeneous swarms. Additionally, the learned interaction forces could be utilized for motion planning and controlling dynamically evolving formations.

\bibliographystyle{IEEEtran}
\bibliography{bib/icra2025}

\begin{thebibliography}{10}
\providecommand{\url}[1]{#1}
\csname url@rmstyle\endcsname
\providecommand{\newblock}{\relax}
\providecommand{\bibinfo}[2]{#2}
\providecommand\BIBentrySTDinterwordspacing{\spaceskip=0pt\relax}
\providecommand\BIBentryALTinterwordstretchfactor{4}
\providecommand\BIBentryALTinterwordspacing{\spaceskip=\fontdimen2\font plus
\BIBentryALTinterwordstretchfactor\fontdimen3\font minus \fontdimen4\font\relax}
\providecommand\BIBforeignlanguage[2]{{%
\expandafter\ifx\csname l@#1\endcsname\relax
\typeout{** WARNING: IEEEtran.bst: No hyphenation pattern has been}%
\typeout{** loaded for the language `#1'. Using the pattern for}%
\typeout{** the default language instead.}%
\else
\language=\csname l@#1\endcsname
\fi
#2}}

\bibitem{abdelkader2021aerial}
M.~Abdelkader, S.~G{\"u}ler, H.~Jaleel, and J.~S. Shamma, ``Aerial swarms: Recent applications and challenges,'' \emph{Current robotics reports}, vol.~2, pp. 309--320, 2021.

\bibitem{chung2018survey}
S.-J. Chung, A.~A. Paranjape, P.~Dames, S.~Shen, and V.~Kumar, ``A survey on aerial swarm robotics,'' \emph{IEEE Transactions on Robotics}, vol.~34, no.~4, pp. 837--855, 2018.

\bibitem{morgan2016swarm}
D.~Morgan, G.~P. Subramanian, S.-J. Chung, and F.~Y. Hadaegh, ``Swarm assignment and trajectory optimization using variable-swarm, distributed auction assignment and sequential convex programming,'' \emph{The International Journal of Robotics Research}, vol.~35, no.~10, pp. 1261--1285, 2016.

\bibitem{danjun2015autonomous}
L.~Danjun, Z.~Yan, S.~Zongying, and L.~Geng, ``Autonomous landing of quadrotor based on ground effect modelling,'' in \emph{2015 34th Chinese control conference (CCC)}.\hskip 1em plus 0.5em minus 0.4em\relax IEEE, 2015, pp. 5647--5652.

\bibitem{kan2019analysis}
X.~Kan, J.~Thomas, H.~Teng, H.~G. Tanner, V.~Kumar, and K.~Karydis, ``Analysis of ground effect for small-scale uavs in forward flight,'' \emph{IEEE Robotics and Automation Letters}, vol.~4, no.~4, pp. 3860--3867, 2019.

\bibitem{honig2018trajectory}
W.~H{\"o}nig, J.~A. Preiss, T.~S. Kumar, G.~S. Sukhatme, and N.~Ayanian, ``Trajectory planning for quadrotor swarms,'' \emph{IEEE Transactions on Robotics}, vol.~34, no.~4, pp. 856--869, 2018.

\bibitem{wensing2023optimization}
P.~M. Wensing, M.~Posa, Y.~Hu, A.~Escande, N.~Mansard, and A.~Del~Prete, ``Optimization-based control for dynamic legged robots,'' \emph{IEEE Transactions on Robotics}, 2023.

\bibitem{di2020software}
J.~Di~Carlo, ``Software and control design for the mit cheetah quadruped robots,'' Ph.D. dissertation, Massachusetts Institute of Technology, 2020.

\bibitem{manchester2019contact}
Z.~Manchester, N.~Doshi, R.~J. Wood, and S.~Kuindersma, ``Contact-implicit trajectory optimization using variational integrators,'' \emph{The International Journal of Robotics Research}, vol.~38, no. 12-13, pp. 1463--1476, 2019.

\bibitem{xu2014gp}
N.~Xu, K.~H. Low, J.~Chen, K.~K. Lim, and E.~Ozgul, ``Gp-localize: Persistent mobile robot localization using online sparse gaussian process observation model,'' in \emph{Proceedings of the AAAI Conference on Artificial Intelligence}, vol.~28, no.~1, 2014.

\bibitem{wilcox2020solar}
B.~Wilcox and M.~C. Yip, ``Solar-gp: Sparse online locally adaptive regression using gaussian processes for bayesian robot model learning and control,'' \emph{IEEE Robotics and Automation Letters}, vol.~5, no.~2, pp. 2832--2839, 2020.

\bibitem{liu2018gaussian}
M.~Liu, G.~Chowdhary, B.~C. Da~Silva, S.-Y. Liu, and J.~P. How, ``Gaussian processes for learning and control: A tutorial with examples,'' \emph{IEEE Control Systems Magazine}, vol.~38, no.~5, pp. 53--86, 2018.

\bibitem{schmid2022real}
N.~Schmid, J.~Gruner, H.~S. Abbas, and P.~Rostalski, ``A real-time gp based mpc for quadcopters with unknown disturbances,'' in \emph{2022 American Control Conference (ACC)}.\hskip 1em plus 0.5em minus 0.4em\relax IEEE, 2022, pp. 2051--2056.

\bibitem{zheng2022gp}
Y.~Zheng, T.~Zhang, S.~Li, G.~Zhang, and Y.~Wang, ``Gp-based mpc with updating tube for safety control of unknown system,'' \emph{Digital Chemical Engineering}, vol.~4, p. 100041, 2022.

\bibitem{nghiem2019linearized}
T.~X. Nghiem, ``Linearized gaussian processes for fast data-driven model predictive control,'' in \emph{2019 American Control Conference (ACC)}.\hskip 1em plus 0.5em minus 0.4em\relax IEEE, 2019, pp. 1629--1634.

\bibitem{nghiem2019fast}
T.~X. Nghiem, T.-D. Nguyen, and V.-A. Le, ``Fast gaussian process based model predictive control with uncertainty propagation,'' in \emph{2019 57th Annual Allerton Conference on Communication, Control, and Computing (Allerton)}.\hskip 1em plus 0.5em minus 0.4em\relax IEEE, 2019, pp. 1052--1059.

\bibitem{nguyen2024tinympc}
K.~Nguyen, S.~Schoedel, A.~Alavilli, B.~Plancher, and Z.~Manchester, ``Tinympc: Model-predictive control on resource-constrained microcontrollers,'' in \emph{2024 IEEE International Conference on Robotics and Automation (ICRA)}.\hskip 1em plus 0.5em minus 0.4em\relax IEEE, 2024, pp. 1--7.

\bibitem{hewing2019cautious}
L.~Hewing, J.~Kabzan, and M.~N. Zeilinger, ``Cautious model predictive control using gaussian process regression,'' \emph{IEEE Transactions on Control Systems Technology}, vol.~28, no.~6, pp. 2736--2743, 2019.

\bibitem{koller2018learning}
T.~Koller, F.~Berkenkamp, M.~Turchetta, and A.~Krause, ``Learning-based model predictive control for safe exploration,'' in \emph{2018 IEEE conference on decision and control (CDC)}.\hskip 1em plus 0.5em minus 0.4em\relax IEEE, 2018, pp. 6059--6066.

\bibitem{kabzan2019learning}
J.~Kabzan, L.~Hewing, A.~Liniger, and M.~N. Zeilinger, ``Learning-based model predictive control for autonomous racing,'' \emph{IEEE Robotics and Automation Letters}, vol.~4, no.~4, pp. 3363--3370, 2019.

\bibitem{morgan2014model}
D.~Morgan, S.-J. Chung, and F.~Y. Hadaegh, ``Model predictive control of swarms of spacecraft using sequential convex programming,'' \emph{Journal of Guidance, Control, and Dynamics}, vol.~37, no.~6, pp. 1725--1740, 2014.

\bibitem{debord2018trajectory}
M.~Debord, W.~H{\"o}nig, and N.~Ayanian, ``Trajectory planning for heterogeneous robot teams,'' in \emph{2018 IEEE/RSJ International Conference on Intelligent Robots and Systems (IROS)}.\hskip 1em plus 0.5em minus 0.4em\relax IEEE, 2018, pp. 7924--7931.

\bibitem{williams2006gaussian}
C.~K. Williams and C.~E. Rasmussen, \emph{Gaussian processes for machine learning}.\hskip 1em plus 0.5em minus 0.4em\relax MIT press Cambridge, MA, 2006, vol.~2, no.~3.

\bibitem{ma2023gaussian}
H.~Ma, T.~Zhang, Y.~Wu, F.~P. Calmon, and N.~Li, ``Gaussian max-value entropy search for multi-agent bayesian optimization,'' in \emph{2023 IEEE/RSJ International Conference on Intelligent Robots and Systems (IROS)}.\hskip 1em plus 0.5em minus 0.4em\relax IEEE, 2023, pp. 10\,028--10\,035.

\bibitem{duvenaud2011additive}
D.~K. Duvenaud, H.~Nickisch, and C.~Rasmussen, ``Additive gaussian processes,'' \emph{Advances in neural information processing systems}, vol.~24, 2011.

\bibitem{kandasamy2015high}
K.~Kandasamy, J.~Schneider, and B.~P{\'o}czos, ``High dimensional bayesian optimisation and bandits via additive models,'' in \emph{International conference on machine learning}.\hskip 1em plus 0.5em minus 0.4em\relax PMLR, 2015, pp. 295--304.

\bibitem{lewis2012optimal}
F.~L. Lewis, D.~Vrabie, and V.~L. Syrmos, \emph{Optimal control}.\hskip 1em plus 0.5em minus 0.4em\relax John Wiley \& Sons, 2012.

\bibitem{kirk2004optimal}
D.~E. Kirk, \emph{Optimal control theory: an introduction}.\hskip 1em plus 0.5em minus 0.4em\relax Courier Corporation, 2004.

\bibitem{todorov2006optimal}
E.~Todorov, ``Optimal control theory,'' 2006.

\bibitem{lu2024mpc}
Y.~Lu, Z.~Li, Y.~Zhou, N.~Li, and Y.~Mo, ``Mpc-inspired reinforcement learning for verifiable model-free control,'' in \emph{6th Annual Learning for Dynamics \& Control Conference}.\hskip 1em plus 0.5em minus 0.4em\relax PMLR, 2024, pp. 399--413.

\bibitem{srinivas2009gaussian}
N.~Srinivas, A.~Krause, S.~M. Kakade, and M.~Seeger, ``Gaussian process optimization in the bandit setting: No regret and experimental design,'' \emph{arXiv preprint arXiv:0912.3995}, 2009.

\bibitem{panerati2021learning}
J.~Panerati, H.~Zheng, S.~Zhou, J.~Xu, A.~Prorok, and A.~P. Schoellig, ``Learning to fly—a gym environment with pybullet physics for reinforcement learning of multi-agent quadcopter control,'' in \emph{2021 IEEE/RSJ International Conference on Intelligent Robots and Systems (IROS)}.\hskip 1em plus 0.5em minus 0.4em\relax IEEE, 2021, pp. 7512--7519.

\bibitem{budaciu2019evaluation}
C.~Budaciu, N.~Botezatu, M.~Kloetzer, and A.~Burlacu, ``On the evaluation of the crazyflie modular quadcopter system,'' in \emph{2019 24th IEEE International Conference on Emerging Technologies and Factory Automation (ETFA)}.\hskip 1em plus 0.5em minus 0.4em\relax IEEE, 2019, pp. 1189--1195.

\bibitem{preiss2017crazyswarm}
J.~A. Preiss, W.~Honig, G.~S. Sukhatme, and N.~Ayanian, ``Crazyswarm: A large nano-quadcopter swarm,'' in \emph{2017 IEEE International Conference on Robotics and Automation (ICRA)}.\hskip 1em plus 0.5em minus 0.4em\relax IEEE, 2017, pp. 3299--3304.

\bibitem{pichierri2023crazychoir}
L.~Pichierri, A.~Testa, and G.~Notarstefano, ``Crazychoir: Flying swarms of crazyflie quadrotors in ros 2,'' \emph{IEEE Robotics and Automation Letters}, vol.~8, no.~8, pp. 4713--4720, 2023.

\end{thebibliography}

\end{document}